\documentclass{article}
\usepackage{stywhispers,amsmath,epsfig}

\usepackage{booktabs} 
\usepackage{tabularx} 

\usepackage[colorlinks=true,linkcolor=black,urlcolor=blue,citecolor=black]{hyperref} 

\usepackage{graphicx} 
\usepackage{subcaption} 
\usepackage[numbers]{natbib} 
\usepackage{amsmath} 
\usepackage{amssymb} 
\usepackage{lipsum} 
\usepackage{xcolor} 

\title{An Open Hyperspectral Dataset with Sea-Land-Cloud Ground-Truth from the HYPSO-1 Satellite}
\name{\begin{tabular}{c}
Jon A. Justo\textsuperscript{1}, 
Joseph Garrett\textsuperscript{1},
Dennis D. Langer\textsuperscript{2}, Marie B. Henriksen\textsuperscript{1}, \\Radu T. Ionescu\textsuperscript{3}, and Tor A. Johansen\textsuperscript{1}
\end{tabular}}

\address{\textsuperscript{1}Dept. Engineering Cybernetics, Norwegian University of Science and Technology, Trondheim, Norway\\
\textsuperscript{2}Dept. Marine Technology, Norwegian University of Science and Technology, Trondheim, Norway\\
\textsuperscript{3}Dept. of Computer Science, University of Bucharest, Romania}

\begin{document}
%
\maketitle
%


\begin{abstract}
Hyperspectral Imaging, employed in satellites for space remote sensing, like HYPSO-1, faces constraints due to few labeled data sets, affecting the training of AI models demanding these ground-truth annotations. 
In this work, we introduce \textit{The HYPSO-1 Sea-Land-Cloud-Labeled Dataset}, an open dataset with 200 diverse hyperspectral images from the HYPSO-1 mission, available in both raw and calibrated forms for scientific research in Earth observation. 
Moreover, 38 of these images from different countries include ground-truth labels at pixel-level totaling about 25 million spectral signatures labeled for sea/land/cloud categories. 
To demonstrate the potential of the dataset and its labeled subset, we have additionally optimized a deep learning model (1D Fully Convolutional Network), achieving superior performance to the current state of the art. 
The complete dataset, ground-truth labels, deep learning model, and software code are openly accessible for download at the website \url{https://ntnu-smallsat-lab.github.io/hypso1_sea_land_clouds_dataset/}.
\end{abstract}




\vspace{-5pt}
\section{Introduction}
Hyperspectral Imaging (HSI) captures and analyzes spectroscopic information from targets of interest by sampling the light they scatter across numerous narrow contiguous spectral channels in the electromagnetic spectrum. 
Unlike conventional RGB and multi-spectral imaging, each spatial pixel in HSI contains a high-resolution spectral signature with multiple channels, enabling HSI to offer valuable insights for the characterization of targets. 
This unique capability has led to widespread applications of HSI for remote sensing for Earth and ocean observation as well as in the medical field \cite{hsi-processing-review-2017}. 

In classification, each spatial pixel is classified into a specific category through the application of e.g.~supervised learning techniques employed in the training of Deep Learning (DL) Neural Networks (NN) where pixel-level ground truth is used to train models to classify the pixels into the labeled categories. 
However, the scarcity of labeled HSI data is evident in the literature of remote sensing, encompassing both Earth and ocean observations. 
Datasets from airborne sensors like AVIRIS \cite{aviris-dataset}, ROSIS \cite{rosis}, and HYDICE \cite{hydice} contain labeled images with diverse land cover categories like urban and agricultural areas. 
Additionally, other popular datasets such as the Kennedy Space Center and Jasper Ridge have limited water coverage due to the capture over small geographic extents using airborne platforms. 
Despite the widespread use of these datasets in HSI classification, each set consists of a single labeled image, inadequate for training emerging data-intensive deep learning models. 
Furthermore, using these airborne platforms for small geographical areas poses challenges in studying broader marine events like Harmful Algal Blooms (HABs) \cite{dickey-bidigare-2005}. 

To meet the demand for more extensive Earth and ocean observations facilitated by HSI technology, satellite platforms are progressively integrating HSI spectrometers. 
For example, the ESA/CHIME mission (Copernicus Hyperspectral Imaging Mission for the Environment) \cite{CHIME_paper} will be utilized for tasks such as characterizing soil properties and managing agriculture and biodiversity. Similarly, the NTNU/HYPSO-1 mission (HYPerspectral small Satellite for Ocean observation) \cite{hypsoconcept} enables oceanography studies by observing ocean colors, facilitating the monitoring and forecast of marine events. 
However, a significant challenge still persists: the lack of easily accessible and labeled HSI satellite imagery for Earth and ocean observations, hindering the training of models that rely on ground-truth annotations and that need to generalize well to unseen data. 

To this end, our focus turns to the HYPSO-1 mission, serving as the space segment of an observational pyramid dedicated to oceanography to study the ocean colors of coastline waters with predominantly aquaculture activities. 
We introduce \textit{The HYPSO-1 Sea-Land-Cloud-Labeled Dataset}, an annotated dataset that additionally marks the first public release of labelled data from the HYPSO-1 satellite, launched on January 13, 2022. 
Our contribution includes: a diverse dataset of 200 images covering all continents with 131 million spectral signatures both in raw and calibrated radiance forms; ground-truth labels for 38 images including sea, land, and cloud categories; a 1D Fully Convolutional Network (FCN) model to demonstrate the potential of our dataset for on-ground and future on-board inference trained for sea/land/cloud classification which is a crucial preprocessing step for upcoming data products like those of HYPSO-1 and CHIME; and the website \url{https://ntnu-smallsat-lab.github.io/hypso1_sea_land_clouds_dataset/} provides openly the resulting dataset, model and Python code. 
Our dataset supports applications like super-resolution, anomaly detection, image fusion, classification, and unmixing.

\vspace{-5pt} 
\section{Methodological Background}
\label{Section:methodology}  

\subsection{Constructing the HYPSO-1 Dataset: 200 Diverse HS Images for Earth and Ocean Observations}
\label{Section:METHODOLOGY_DATASET_CONTRUCTION}
For data collection, HYPSO-1 utilizes a push-broom scanning technique, employing a sensor array with a 684-pixel slit \cite{Bakken2023, elizabethHSICots2021}. 
As the satellite orbits, the spectrometer slit typically scans 956 line frames sequentially. The spatial resolution in the across- and along-track directions is approximately 100 m x 600 m, respectively ~\cite{hypsoconcept,Bakken2023}, and it measures the light across 120 channels with a spectral resolution of roughly 5 nm across the visible and part of the Near-Infrared (NIR) spectra from 400 to 800 nm.
The resulting image is subsequently stored as a data cube with dimensions of $956\times684\times120$. 

Since its launch in January 2022, HYPSO-1 has gathered over a thousand of these HS images. 
The focus of our work lies on the images captured between June and December 2022 in addition to a few captures from February and March 2023. During the image selection process, we adhere to the following criteria. 
First, we concentrate on captures that contain the necessary metadata to perform radiometric calibration. Additionally, we prioritize diverse HS images that represent various scenes. 
We concentrate on scenes with large water bodies, e.g.~found in coastal regions with beaches, fjords and estuaries. 
These scenes display varying water colors, ranging from darker blues to lighter turquoise and green, influenced by reflections from the seabed and algal biomass. 
We include additionally images of inland water bodies, such as large rivers and lakes. 
The HS images in the resulting dataset depict a broad range of landscapes, with varying cloud density and light exposure parameters. 
While our emphasis lies on water bodies within coastal and marine environments, our dataset also includes forests, arid landscapes with limited vegetation, polar and glacial regions with ice and snow, urban landscapes, and cloud-dominated scenes.


\subsection{Dataset Calibration: From Digital Number (Raw) to Radiance}
\label{Section:METHODOLOGY_DATASET_CALIBRATION}
We calibrate raw sensor units to physical values as per~\cite{Henriksen2022b}. Spectral calibration assigns accurate wavelengths to channels using a second-order polynomial fit~\cite{Henriksen2022c}, and adds the wavelengths as metadata.  
Secondly, we employ radiometric calibration to convert from Digital Number (DN) to radiance using the original radiometric coefficients from~\cite{Henriksen2022b}. 
Stripes and other artefacts reported in \cite{Bakken2023,Henriksen2022d} are not yet accounted for. Additionally, smile and keystone corrections are not considered here, although they have been characterized and can be rectified as detailed in~\cite{Henriksen2022b}. Such corrections will undergo further testing before integration into the calibration pipeline. Finally, the calibration does not apply to overexposed pixels, as the correct intensity is unknown when a pixel is saturated, preventing conversion to a known radiance value.


\subsection{Ground-Truth Labels: Semi-Automatic Pixel-Level Annotation for Sea-Land-Clouds Classification}
\label{Section:METHODOLOGY_DATASET_LABELING}
Using the ENVI (version 5.6.2, L3Harris Geospatial) segmentation tool, we perform semi-automatic pixel-level labeling for each of the 38 images selected as a subset from our dataset of 200 images. 
This set of images is carefully selected to cover various scenes. 
Polar and glacial regions are deliberately excluded to avoid mixing up clouds with ice or snow. 
Next, we propose a semi-automatic labeling process designed for precise results achieved relatively fast.

We start by categorizing key classes: extensive water bodies as ``sea'', land covers under ``land'' and targets such as clouds and overexposed pixels under ``clouds'' as these later pixels often obscure the actual targets of interest. 
Our labeling process is initiated through the establishment of training areas for each of these classes. 
This involves the manual extraction of data features via Regions of Interest (ROIs) defined by drawing polygons for each class on an RGB composite generated from the raw HS image. 
Each polygon becomes a training area for ENVI's Machine Learning (ML) models that subsequently predict the categories for the full image, including non-polygon pixels. 
Precise polygon drawing, especially at class boundaries, is crucial due to its significance ensuring reliable class labeling. 

After polygon creation, we follow the subsequent iterative process that is essential for accurate labeling. 
First, we pick one of the four ML models available in the tool: Maximum Likelihood, Minimum Distance, Mahalanobis Distance, or Spectral Angle Mapper. 
The selected algorithm learns from the training ROIs and classifies pixels based on feature similarity. 
For instance, the Mahalanobis Distance predicts the category of each pixel by measuring the distance between its feature vector and the average signature associated with the corresponding class. 
The ML model's effectiveness relies on the quality of the training areas and the inherent characteristics of the image under training. 
Second, if the algorithm does not yield the desired labeling results, an alternative, instead of changing the ML algorithm itself, is to fine-tune the polygons. 
This includes adding more diverse polygons, especially along class borders, and reviewing statistical spectral information for each ROI.

Finally, we refine the labeling results using techniques such as smoothing and aggregation to correct anomalies in the labels. 
When necessary, a team of experts manually corrects the annotations to ensure that the quality of the ground-truth labels matches the human-level expertise in the field.


\begin{figure*}[ht] 
    \centering
    \newcommand{\subfigwidth}{0.25\columnwidth} 

    \begin{subfigure}[b]{\subfigwidth}
        \begin{minipage}{\textwidth}
            \includegraphics[width=\linewidth]{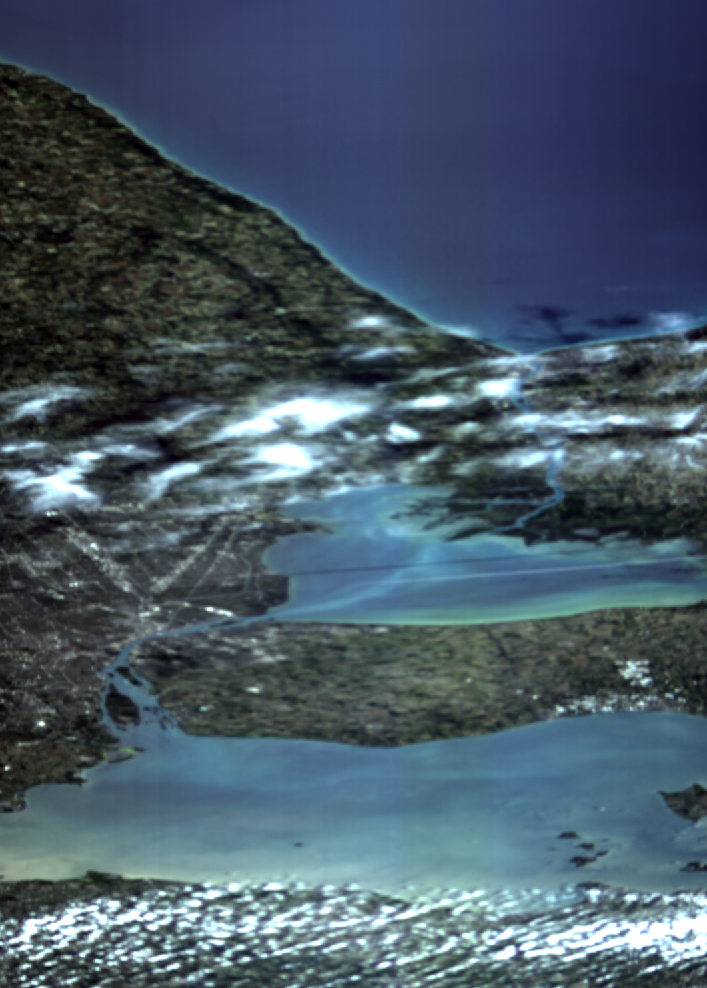}\hfill\includegraphics[width=\linewidth]{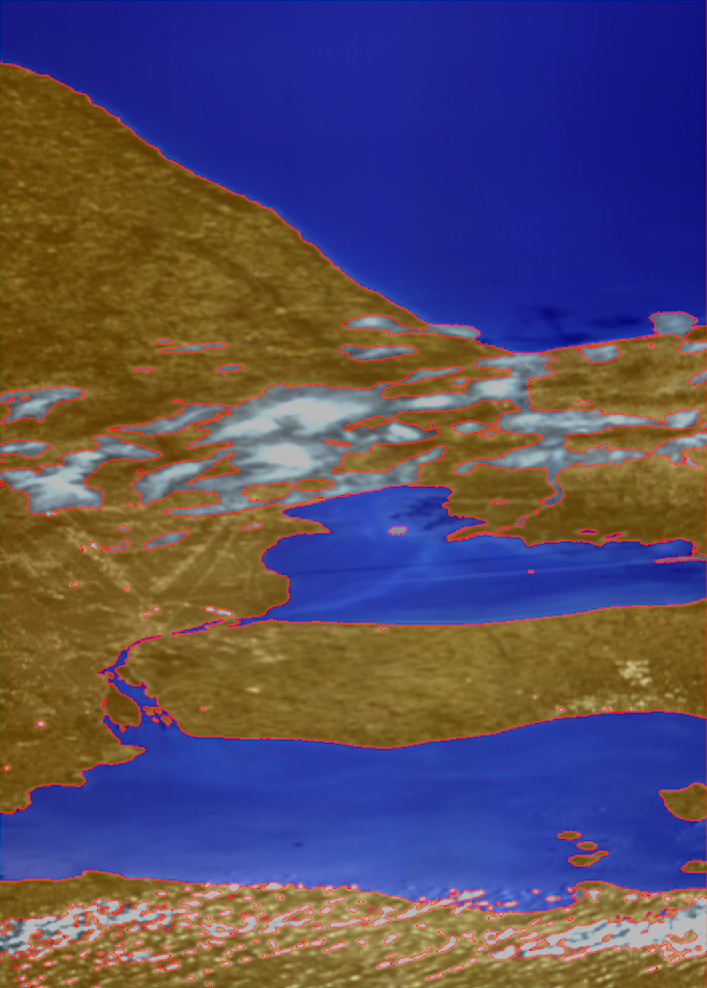}
        \end{minipage}
        \caption{}
        \label{Fig:253_ERIE_NO_URBAN}
    \end{subfigure}
    \begin{subfigure}[b]{\subfigwidth}
        \begin{minipage}{\textwidth}
            \includegraphics[width=\linewidth]{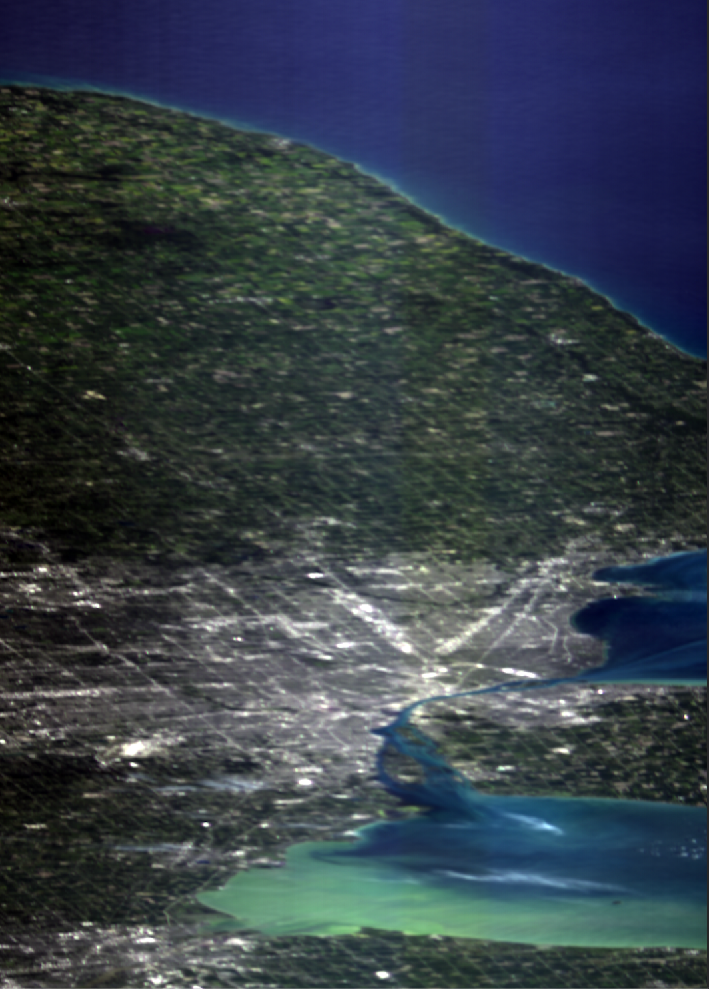}\hfill\includegraphics[width=\linewidth]{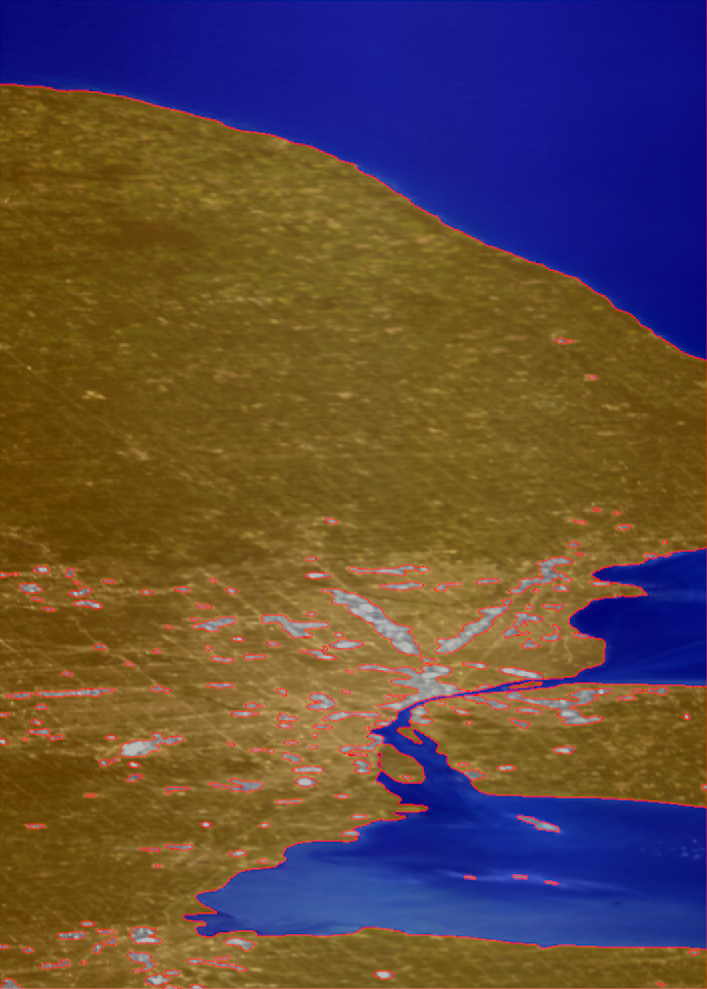}
        \end{minipage}
        \caption{}
        \label{Fig:17_ERIE_URBAN}
    \end{subfigure}
    \begin{subfigure}[b]{\subfigwidth}
        \includegraphics[width=\textwidth]{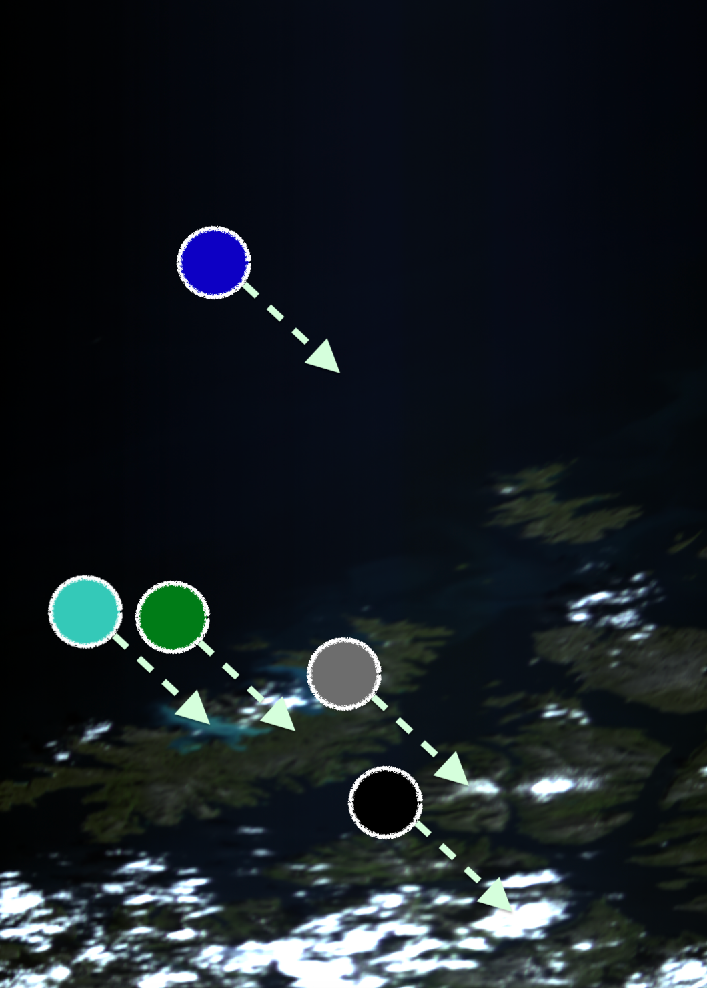}\hfill\includegraphics[width=\textwidth]{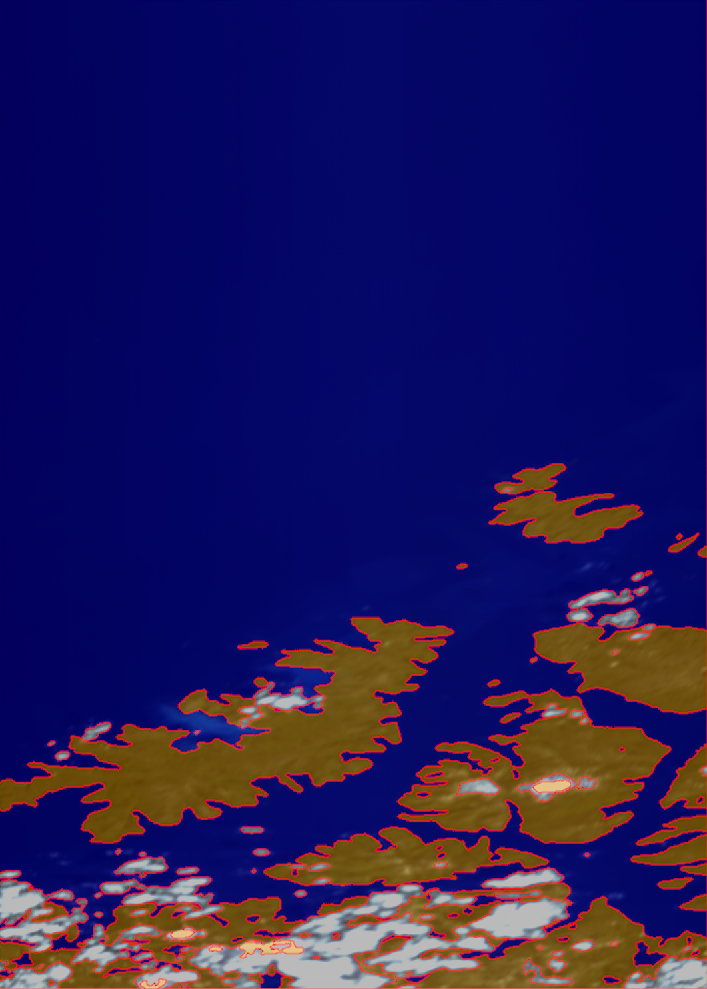}
        \caption{}
        \label{Fig:211_FINNMARK}
    \end{subfigure}
    \begin{subfigure}[b]{\subfigwidth}
        \includegraphics[width=\textwidth]{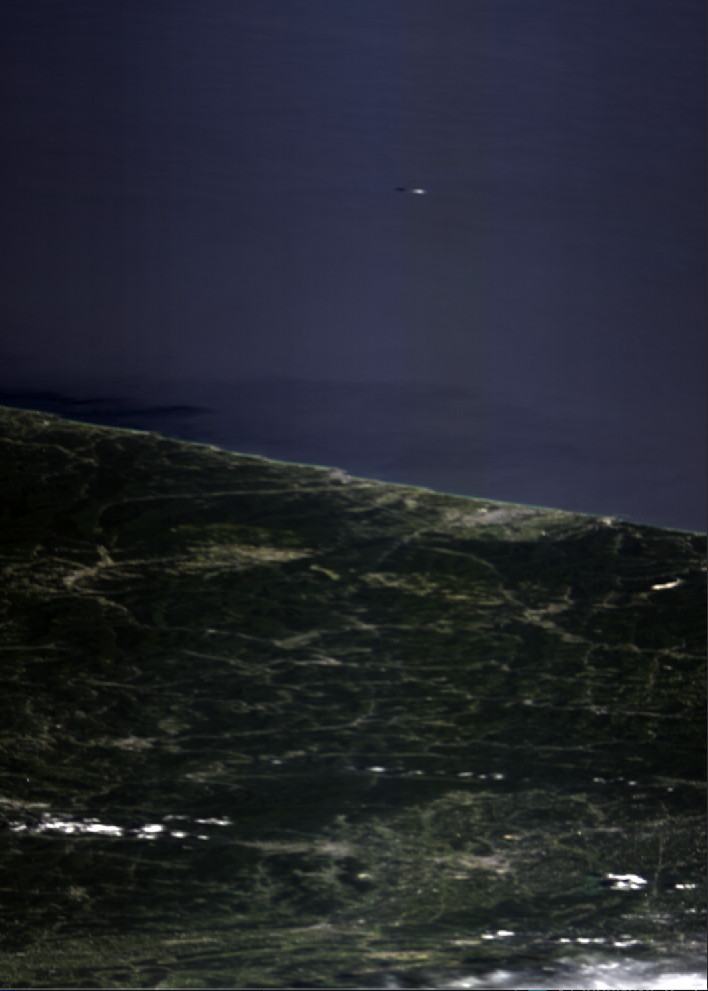}\hfill\includegraphics[width=\textwidth]{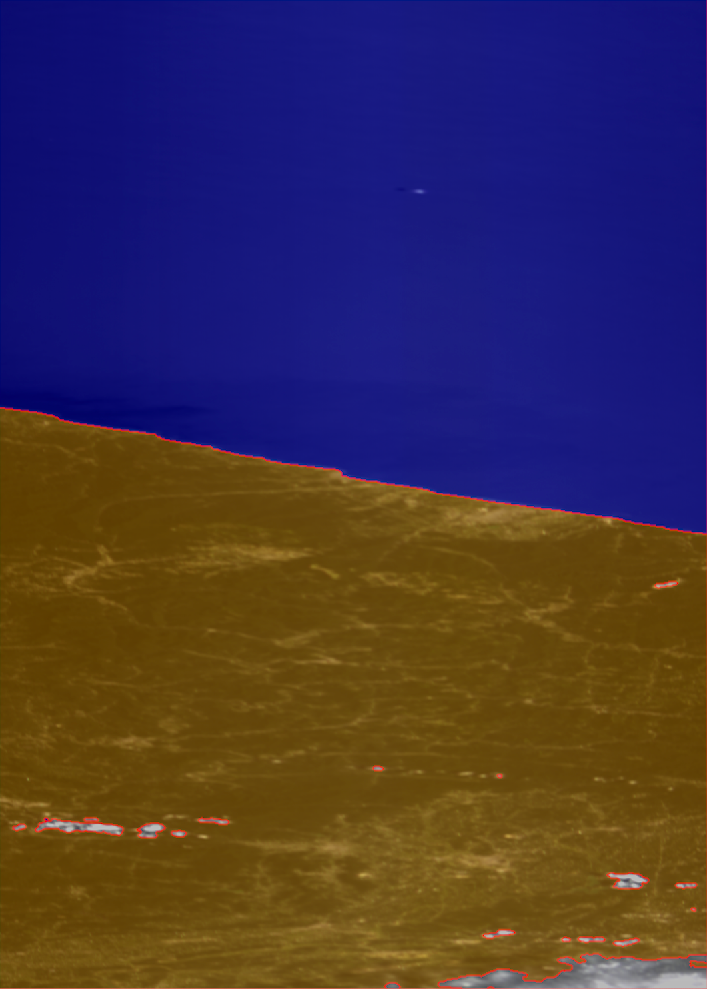}
        \caption{}
        \label{Fig:4_INCHEON}
    \end{subfigure}    
    \begin{subfigure}[b]{\subfigwidth}
        \includegraphics[width=\textwidth]{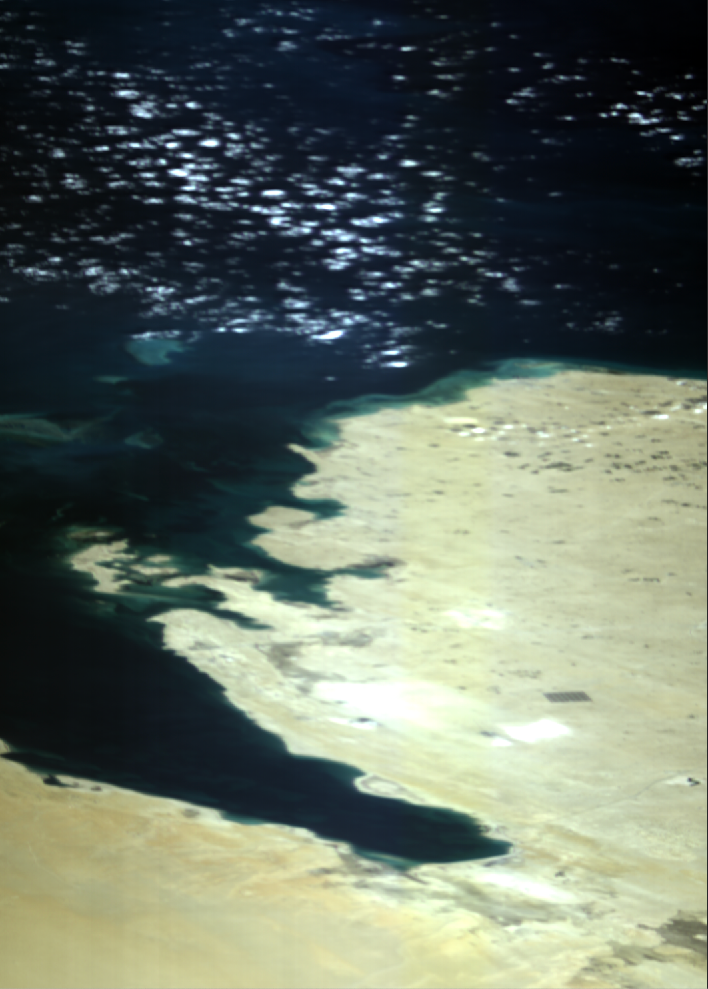}\hfill\includegraphics[width=\textwidth]{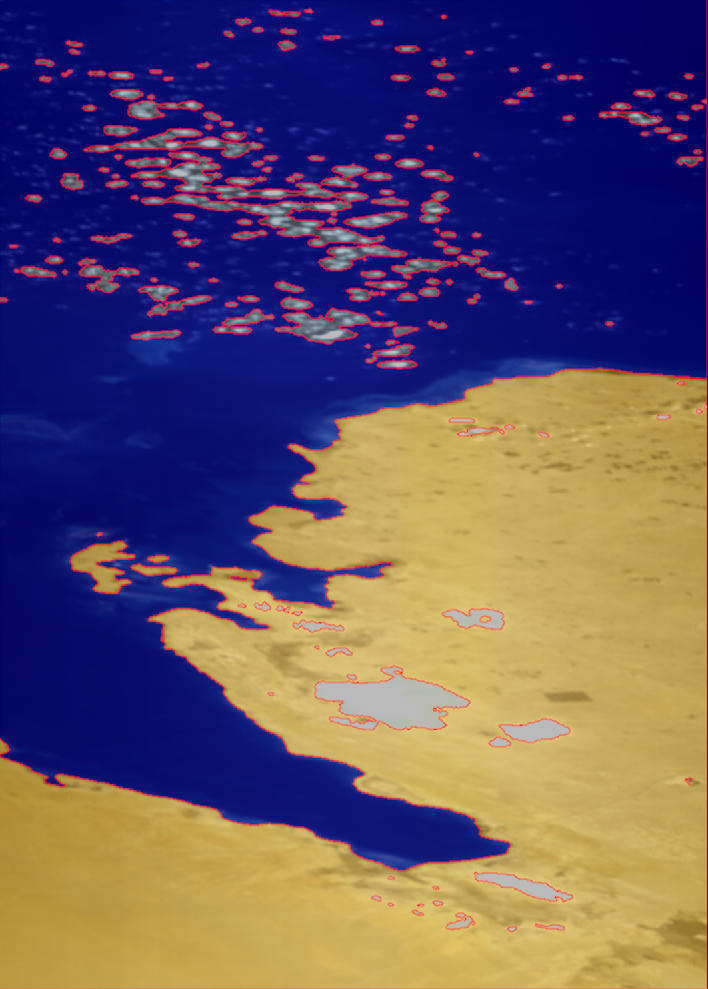}
        \caption{}
        \label{Fig:39_QATAR}
    \end{subfigure}  
    \begin{subfigure}[b]{\subfigwidth}
        \includegraphics[width=\textwidth]{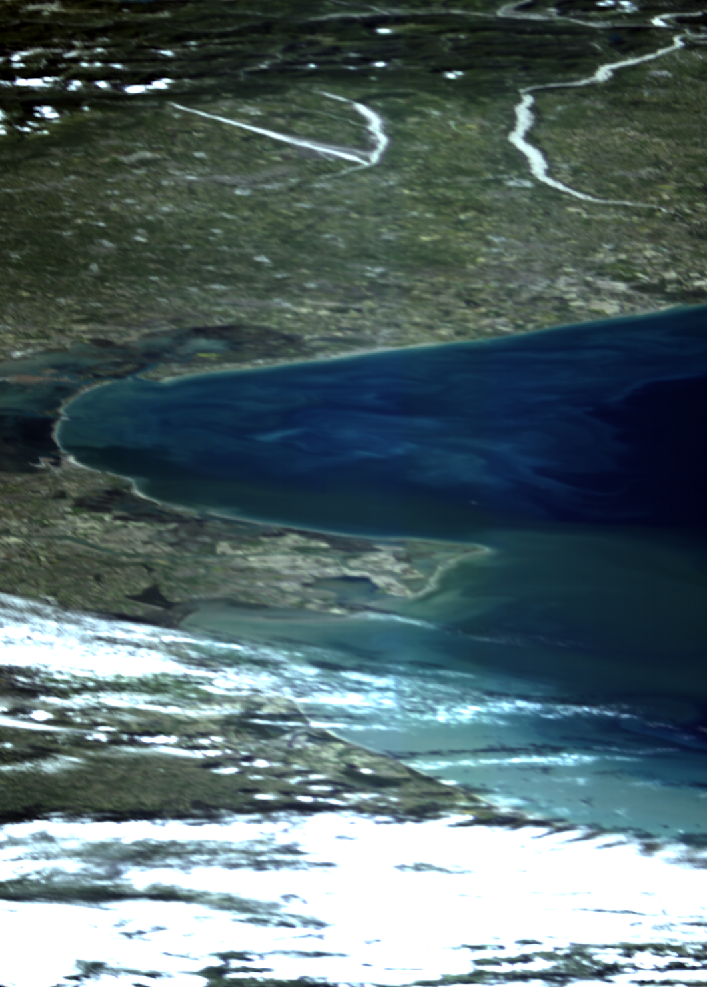}\hfill\includegraphics[width=\textwidth]{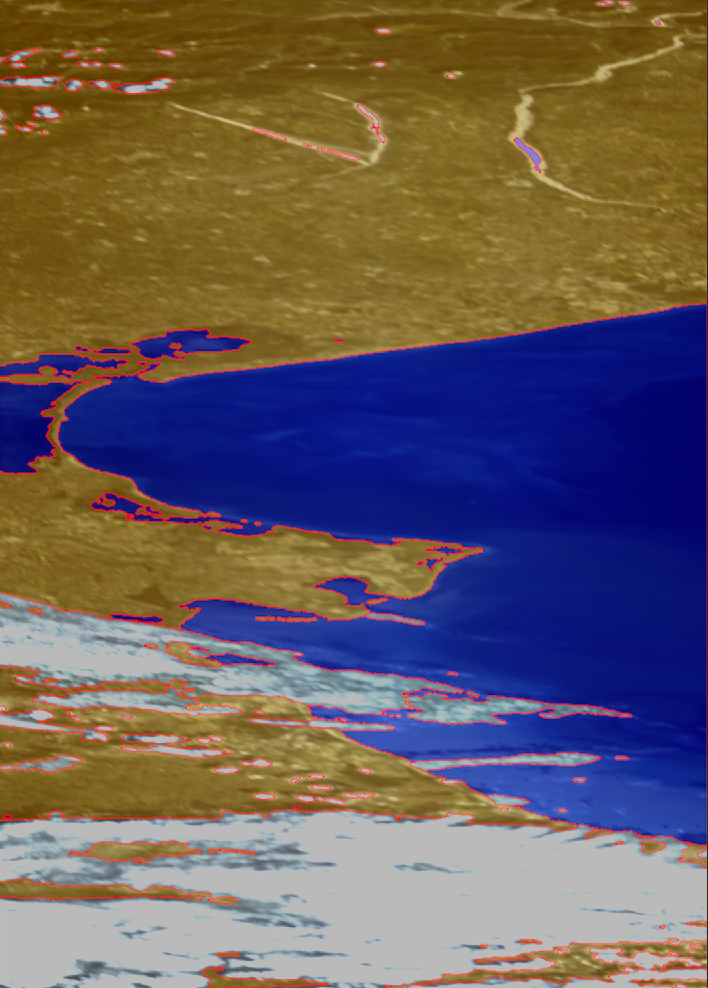}
        \caption{}
        \label{Fig:15_VENICE}
    \end{subfigure}
    \begin{subfigure}[b]{\subfigwidth}
        \includegraphics[width=\textwidth]{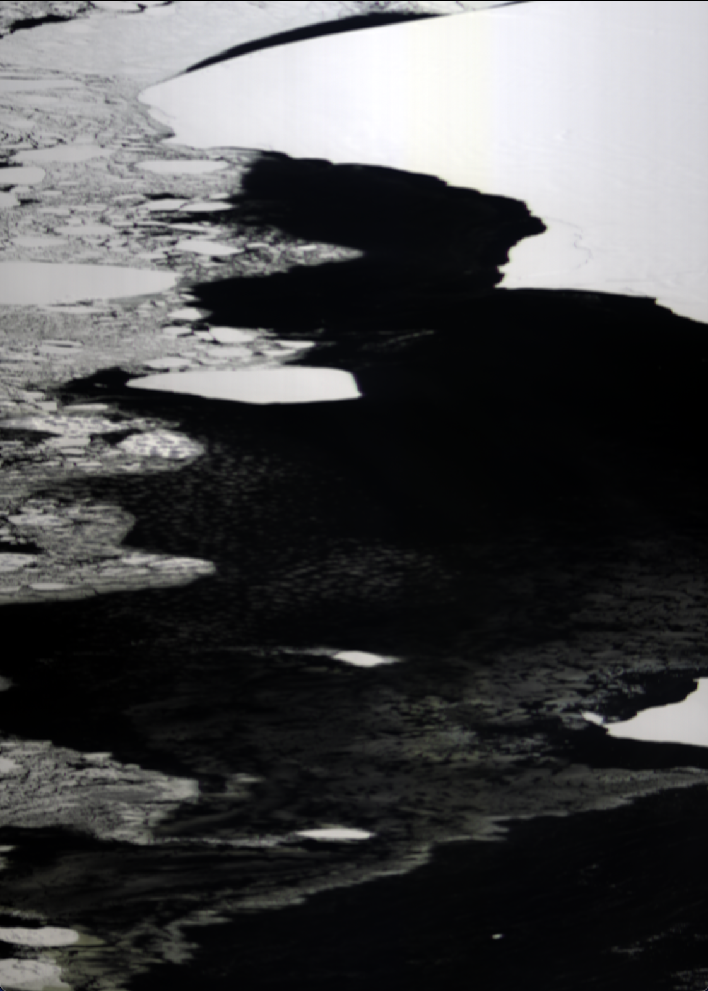}\hfill
        \includegraphics[width=\textwidth]{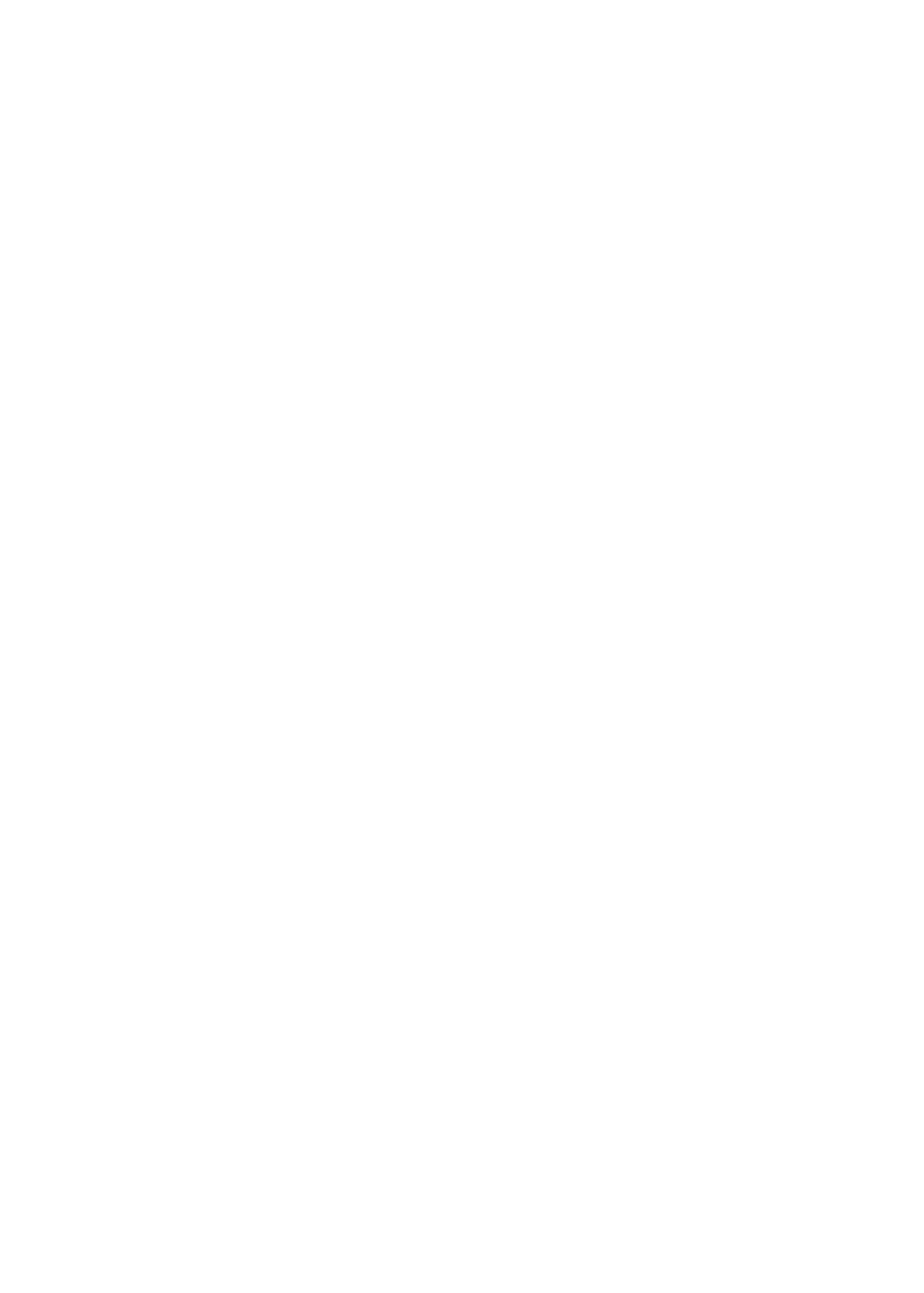}
        \caption{}
        \label{Fig:53_PENGUIN}
    \end{subfigure}

    \caption{RGB composites with ground-truth labels: (blue: ``sea'', orange: ``land'', and gray: ``clouds''). (\ref{Fig:253_ERIE_NO_URBAN}) and (\ref{Fig:17_ERIE_URBAN}): Lake Erie situated between the USA and Canada on 2022/07/09 and 2022/08/27. (\ref{Fig:211_FINNMARK}): Finnmark in Norway on 2022/08/06. (\ref{Fig:4_INCHEON}): South Korea on 2022/08/27. (\ref{Fig:39_QATAR}): West of Qatar on 2022/12/13. (\ref{Fig:15_VENICE}): Venice in Italy on 2022/09/21. (\ref{Fig:53_PENGUIN}): Antarctica on 2022/12/06.}
    \label{Fig:DATA_CAPTURES_FIGURE}
\end{figure*}

\subsection{Example of a Supervised Deep Learning Model: Pixel-Level Sea-Land-Clouds Classification by a 1D FCN}
\label{Section:METHODOLOGY_FCN}

\begin{figure}[htbp]
    \centering 
\includegraphics[width=0.45\textwidth]{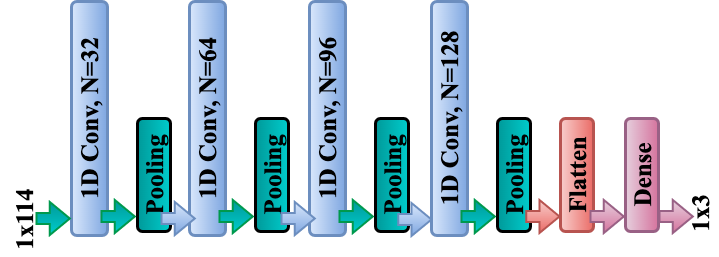} 
    \caption{Adjusted 1D FCN Architecture for Sea-Land-Clouds Supervised Classification for HYPSO-1 HS Images.}
    \label{Fig:Liutetal_architecture} 
\end{figure}

To demonstrate the significance of our dataset for Earth and ocean observation, HSI data processing and sea/land/clouds classification tasks, we adapt the following 1D FCN architecture to tackle this classification problem. 
The 1D FCN, originally designed for regression-based predictions of e.g. clay content in soil spectroscopy, as detailed in \cite{liuetal2018a_original_one_dimensional} (2018) was later adapted to address classification tasks for soil texture as per \cite{soil_texture_one_dimensional_CNN_2019} (2019). 
In our work, as depicted in Fig.~\ref{Fig:Liutetal_architecture}, we employ the 2019 architecture from \cite{soil_texture_one_dimensional_CNN_2019}, but we make several adjustments to its hyper-parameters to boost its performance on our dataset. 
Contrary to the 2019 classification model which utilizes four convolutions with 32, 32, 64, and 64 kernels (all of size 3), our convolutions are expanded to 32, 64, 96, and 128 kernels, increasing the kernel size to 6 for each filter to enhance our neural network's efficacy. 
Despite using more and larger kernels, our model, with 124,163 parameters trained using categorical cross-entropy loss over one epoch in 300-pixel batches on raw data excluding the first and last three noisy channels, remains relatively simple. This suggests the model is apt for inference, suitable both for on-ground processing and for the System-on-Chip (SoC) software on the HYPSO-1 platform \cite{rs15153756}.

\vspace{-15pt}
\section{Results}
\label{Section:Results}

We have developed a website, accessible at \url{https://ntnu-smallsat-lab.github.io/hypso1_sea_land_clouds_dataset/} 
where we offer further details as well as the full dataset and an overview of the images, ground-truth labels, 1D FCN model, and Python code facilitating intuitive access for researchers.

\vspace{-5pt}
\subsection{The HYPSO-1 Sea-Land-Cloud-Labeled Dataset} 
\label{Section:RESULTS_DATA}
Fig.~\ref{Fig:DATA_CAPTURES_FIGURE} displays a diverse range of raw labeled images in its top row, while the bottom row overlays labels—blue for water systems, orange for land cover, gray for cloudy/overexposed regions, and red for the class boundaries. 
Fig.~\ref{Fig:253_ERIE_NO_URBAN} and \ref{Fig:17_ERIE_URBAN} depict Lake Erie's coastal region, between the USA and Canada, known for freshwater issues like HAB events and pollution. 
Fig.~\ref{Fig:211_FINNMARK} captures Finnmark, Norway, a hub for farming Atlantic Salmon facing also challenges such as HAB events leading to salmon suffocation due to oxygen depletion, emphasizing the importance of monitoring these abundant seafood habitats. The figure includes multiple circles pointing to distinct areas for the analysis of spectral signatures that will be elaborated in the subsequent section. Fig.\ref{Fig:4_INCHEON} shows the cloud-free East coast of South Korea. Fig.\ref{Fig:39_QATAR} examines Qatar's western region. Fig.\ref{Fig:15_VENICE} depicts Venice, Italy. Beyond these geographical regions, the dataset also encompasses a collection of 15 polar images. Noteworthy among them is Fig.\ref{Fig:53_PENGUIN} (unlabeled) of Antarctica's ice, crucial for understanding climate change in polar areas. The labeling process for the presented images in these locations mainly employs the ML Maximum Likelihood Algorithm, except for Fig.~\ref{Fig:17_ERIE_URBAN} and ~\ref{Fig:4_INCHEON}, where the Mahalanobis Distance model is used due to its higher accuracy. 
Among the 38 labeled images, the class distribution comprises 37.01\% water, 40.14\% land, and 22.85\%  clouds/overexposed pixels, resulting in a total of approximately 25 million labeled spectral signatures. 



\vspace{-5pt}
\subsection{Analysis of Spectral Signatures of Different Targets in Finnmark, Norway}
\begin{figure}[th] 
    \centering
    \newcommand{\subfigwidth}{1.0\columnwidth} 
    \begin{subfigure}[b]{\subfigwidth}
        \includegraphics[width=\linewidth]{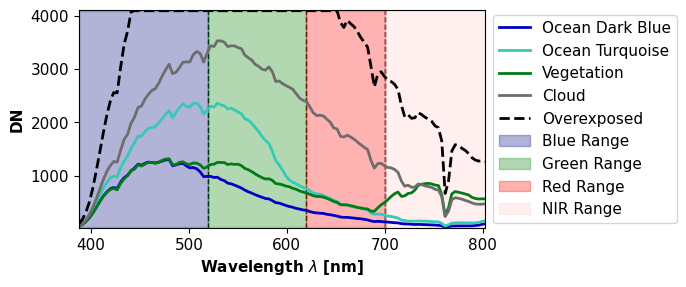}
        \caption{Raw Spectral Signatures: Binned x9 and Reversed to Increasing Wavelength}
        \label{Fig:spectral_signarures_raw}
    \end{subfigure} \hfill
    \begin{subfigure}[b]{\subfigwidth}
        \includegraphics[width=\linewidth]{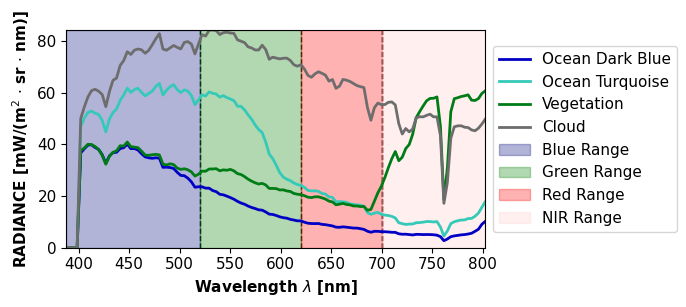}
        \caption{Calibrated Spectral Signatures}
        \label{Fig:spectral_signarures_calibrated}
    \end{subfigure}
    \caption{Signatures from Finnmark (Norway) on 6th August, 2022 (Fig.~\ref{Fig:211_FINNMARK}).}
    \label{Fig:spectral_signarures_ALL}
\end{figure}

We select pixels representing diverse targets in the scene of Finnmark, Norway on 6th August, 2022 (Fig.~\ref{Fig:211_FINNMARK}). 
A dark blue ocean region (blue circle), turquoise ocean color near the coastline (turquoise circle), and vegetated land cover (orange circle) are highlighted. Additionally, a gray circle points to a non-saturated cloud, while a black circle signals a saturated pixel.
Their spectral characteristics are examined in both raw and calibrated formats, as illustrated in Fig.~\ref{Fig:spectral_signarures_ALL}.  
For the overexposed pixel, only the raw signature is displayed, since the calibrated values are not valid. 

Fig.~\ref{Fig:spectral_signarures_ALL} shows that the ocean signatures absorb light in NIR and visible red, as expected. 
Blue and green light is reflected, particularly in shallow coastal waters where the turquoise shade arises from increased visible green reflection from the seafloor and due to factors like suspended material or algal phytoplankton. 
Furthermore, the vegetation signature exhibits a characteristic ``red edge'' effect seen in healthy plants where vegetation poorly absorbs NIR light, since chlorophyll's primary absorption is focused on visible red light for photosynthesis. Visible green also displays notable reflection, contributing to the vegetation's green appearance. For the cloud spectral signature, strong reflectance can be seen for most wavelengths in the visible spectrum. This is as expected from clouds, even though the exact spectral signature may vary with factors like water droplet size.

High contrast scenes can lead to oversaturation in the data due to the sensor's limited dynamic range, especially in settings like cloudy coastal areas or near deserts while recording with an exposure time suitable to image the ocean. Fig.~\ref{Fig:spectral_signarures_raw} shows a saturated pixel with a flat appearance, indicating that the sensor's max light threshold has been reached.


\begin{figure}[htb]
    \centering \includegraphics[width=0.519\textwidth]{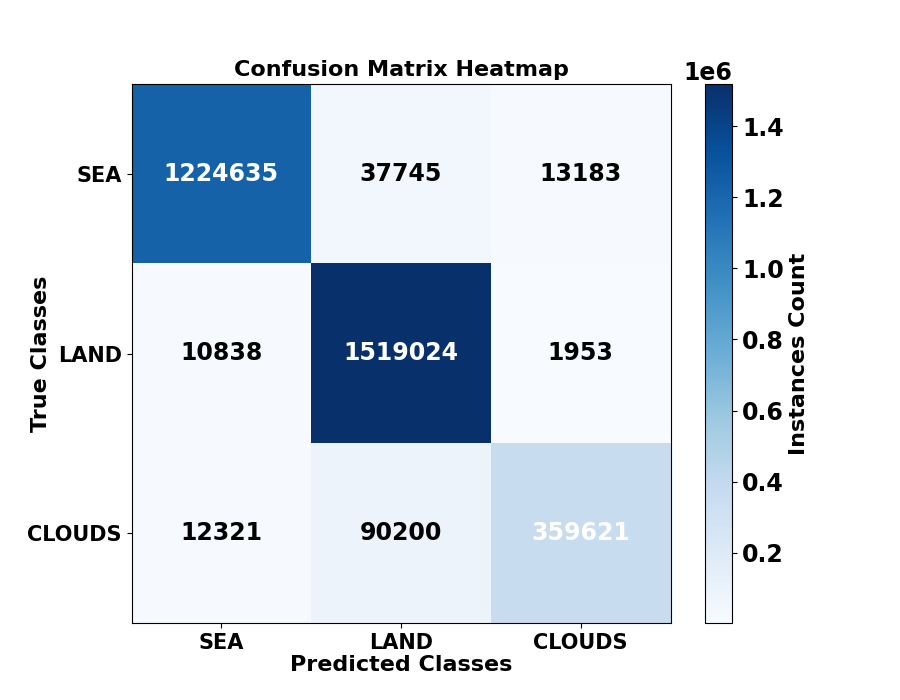}

    \includegraphics[width=0.1215\textwidth]{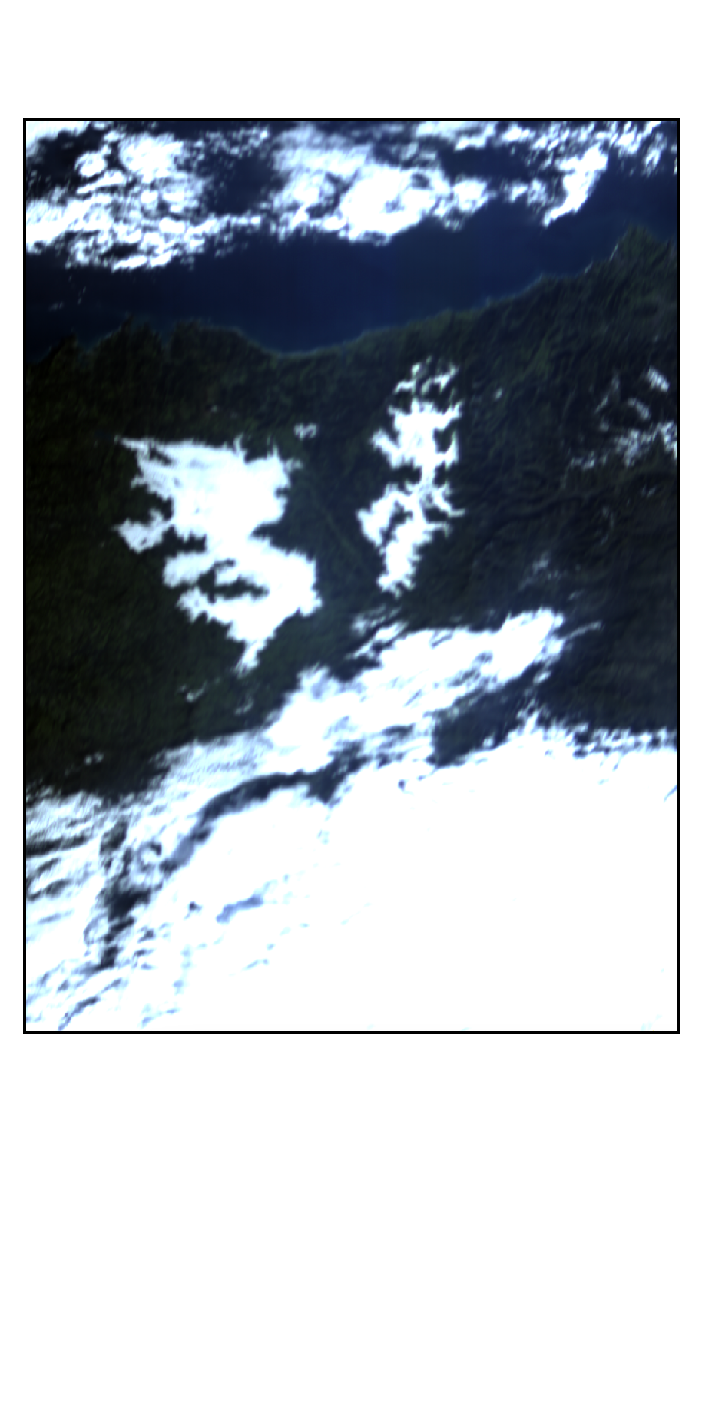}
    \includegraphics[width=0.17\textwidth]{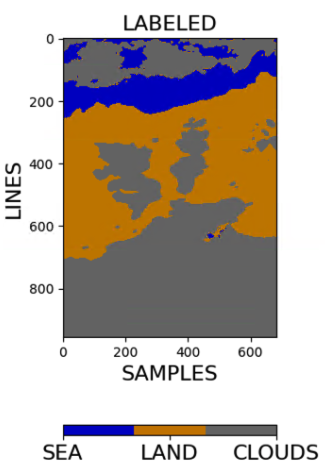} \includegraphics[width=0.17\textwidth]{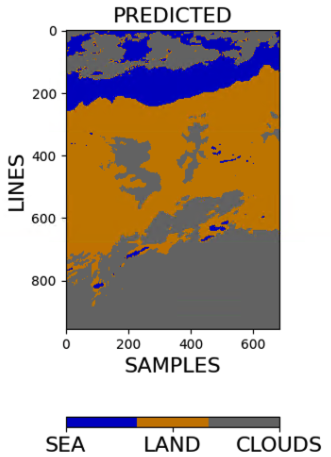}

    \caption{1D FCN Performance on the Test Set. Illustrative Scene: Estuary Ría de Arousa in Galicia, Spain on 4th December, 2022.}
    \label{Fig:Prediction_Testing_Image} 
\end{figure}

\vspace{-5pt}
\subsection{Performance Evaluation of the 1D FCN Model for Pixel-Level Sea-Land-Clouds Classification}
\label{Section:RESULTS_FCN_model}

The training, evaluation and testing stages for the 1D FCN employ the 38 labeled HYPSO-1 images, which are divided as follows: 30 images are used for training (around 20 million signatures), 3 images for validation (approximately 2 million signatures), and 5 images for testing (over 3 million signatures). 
Each spectral signature, with nearly 120 color features, is treated as a one-dimensional data point. 
Our adapted 1D FCN demonstrates a significantly improved performance when compared to the reported results in the literature of HSI-spectroscopy for soil characterization. 
In particular, according to Table 2 in~\cite{soil_texture_one_dimensional_CNN_2019}, the state-of-the-art 1D FCN model achieves an overall accuracy of 0.70, average accuracy of 0.59, and kappa of 0.54, on the LUCAS Dataset. 
In contrast, our adapted 1D FCN model applied to the HYPSO-1 Sea-Land-Cloud-Labeled Dataset achieves substantially higher scores: 0.95 for overall accuracy, 0.91 for average accuracy, and 0.92 for kappa. 
These results highlight the superior performance of our network configuration for our dataset compared to the existing literature. 

Furthermore, Fig.~\ref{Fig:Prediction_Testing_Image} illustrates the model's performance via the confusion matrix for predictions across the entire test set. The calculated accuracy hit rate from the confusion matrix is 0.96 for the ``sea'' class, 0.99 accuracy for the ``land'' class, and 0.78 for the ``clouds'' class, indicating that the ``clouds'' class demonstrates a comparatively lower hit rate accuracy, which we attribute to the dataset's slight skewed distribution, where clouds and overexposed pixels constitute a minority class. 
Moreover, the figure illustrates the pixel-level classification of a HYPSO-1 image from the test set captured in the Estuary Ría de Arousa in the region of Galicia, Spain, on December 4th, 2022 where we utilize dark blue to represent sea water, orange for land, and gray for clouds. The figure presents first the RGB composite, followed by the corresponding ground-truth labeled image and the subsequent 1D FCN model's prediction. 
The model's performance is notably poorer at class boundaries and in identifying clouds and overexposed areas, consistently with the previous results on the test set.

\vspace{-5pt}
\section{Conclusion and Further Work}
\label{Section: Conclusion}
With limited labeled hyperspectral imaging data for Earth and ocean observation, we have introduced a diverse dataset of 200 images, including pixel-level labels for 38 images with sea/land/cloud categories. Using our dataset, we have effectively trained a 1D FCN deep learning model for sea/land/cloud classification, achieving superior performance compared to the state of the art. Future research includes onboard inference for HYPSO-1 and a range of HSI processing tasks like super-resolution, anomaly detection, and image fusion. The dataset, labels, 1D FCN, and Python sample codes are openly available at \url{https://ntnu-smallsat-lab.github.io/hypso1_sea_land_clouds_dataset/}.



\label{sec:ref}

\vspace{-5pt} 

\section*{Acknowledgements}

This research is funded by the NO Grants 2014 – 2021 under Project ELO-Hyp contract no. 24/2020 and Research Council of Norway grants no. 223254 (AMOS), 328724 (Green Platform), and 325961 (HYPSCI). Special thanks to I. Georgescu, A. Masood, and the NTNU Small Satellites team.

\vspace{-5pt} 
\bibliographystyle{IEEEbib-abbrev-first-names-add-etal} 
\bibliography{strings,refs}

\end{document}